\documentclass[letterpaper, 10pt, conference]{ieeeconf}
\IEEEoverridecommandlockouts   
\usepackage{cite}
\usepackage[T1]{fontenc}
\usepackage[utf8]{inputenc}
\usepackage{booktabs,multirow,array}
\usepackage{amsmath,amssymb}
\usepackage{xcolor}
\usepackage{graphicx}
\usepackage{dblfloatfix}
\usepackage{microtype}
\usepackage{caption}
\newcommand{\ci}[3]{$#1$\,[$#2$,\,$#3$]}
\usepackage[hidelinks]{hyperref}

\begin{document}
\title{\textbf{CUSP}: \textbf{CU}SUM-Governed \textbf{S}urvival Hazard Alarms\\ at the \textbf{P}erception Onset for Off-Road Navigation}

\author{Inuk Kang and Seung-Woo Seo (Seoul National University)}

\maketitle
\begin{abstract}
Off-road navigation exposes a robot to potentially hazardous terrain en route.
Although learning-based navigation uses safety
supervision to choose which path to drive, it provides no runtime
alarm when the robot following that path is heading into danger.
Such an alarm must be learned from field logs, where human
intervention preempts the failure and the failure itself is therefore
never observed. The human judges driving unsafe early but typically
intervenes only once failure is clearly near, so the intervention
marks that judgment late. That earlier judgment is what a runtime alarm
must detect, yet no prior intervention-supervised method has targeted
it. To address this problem, we introduce \textbf{CUSP}
(\textbf{CU}SUM-governed \textbf{S}urvival model of the
\textbf{P}erception onset), a model-agnostic runtime hazard alarm
that learns this moment from intervention-terminated logs.
\emph{``Cusp''} is a word for the point at which one state is about to
turn into another, and the moment we target is exactly such a cusp:
the point at which safe driving turns unsafe in a human's judgment.
We call this point the \emph{perception onset} and annotate it separately
from the intervention. A visual hazard head is trained on the annotated
onset with a discrete-time survival objective so that driving with and
without an onset both supervise the head, and a CUSUM accumulates the
predicted onset risk into alarms. We evaluate CUSP at five unseen sites, two
autonomous and three teleoperated, with 142 events and every method
tuned to the same rate of ten false alarms per hour. CUSP detected 85
events compared to 26 for the best of nine adapted baselines, and the
margin comes from hazards to which every signal in the navigation
model is blind.
\end{abstract}

\begin{figure}[t]
\centering
\includegraphics[width=\columnwidth]{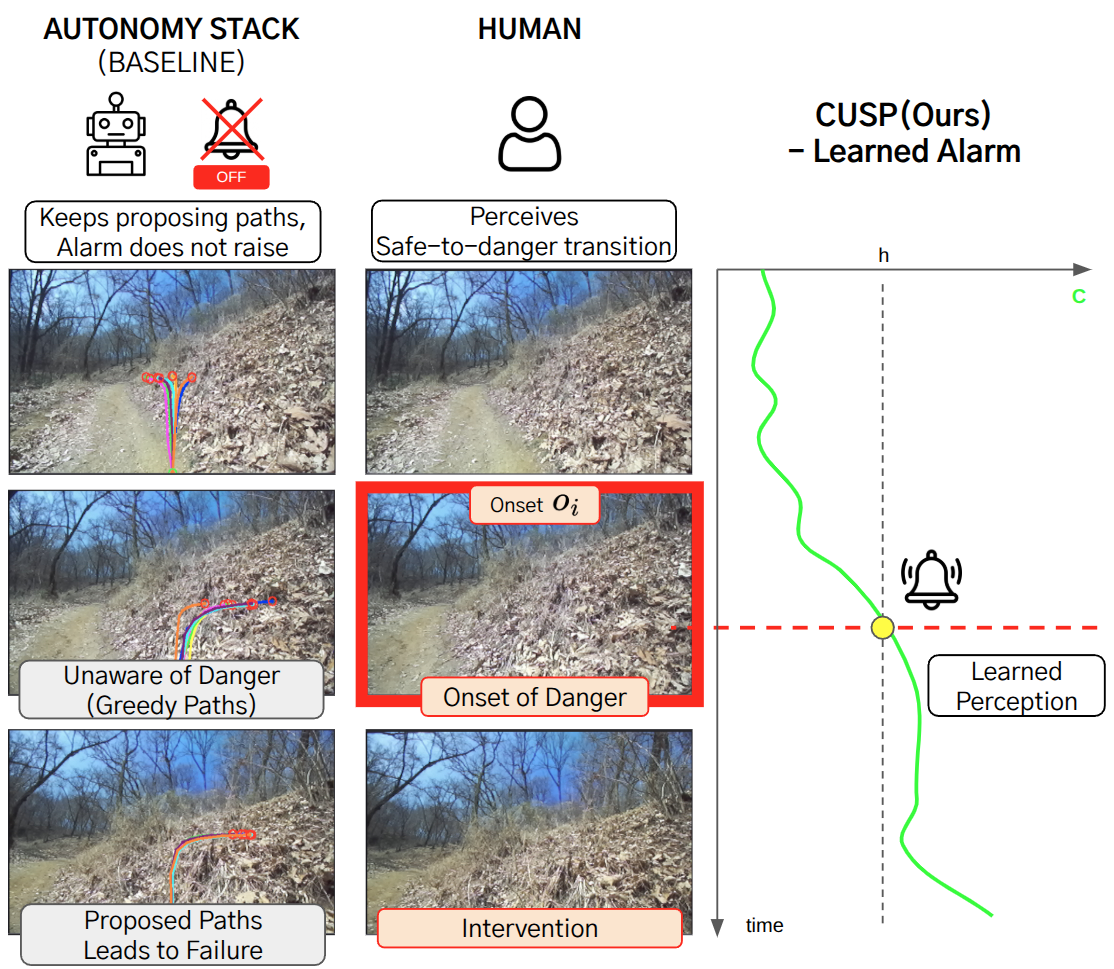}
\caption{The challenge this paper addresses. \textbf{(Left)} The
autonomy model keeps proposing candidate paths as the robot
approaches a deteriorating slope, and nothing in the model raises an
alarm. \textbf{(Middle)} The human perceives the safe-to-dangerous
transition, marked as the onset $o_i$ (red box), and takes over
before failure. \textbf{(Right)} CUSP learns this perception. It
accumulates evidence of the transition over time and raises an alarm
once that evidence exceeds a threshold, here shortly after the onset.}
\label{fig:motivation}
\end{figure}

\section{Introduction}
\label{sec:introduction}

The safety issue in learning-based off-road navigation is handled
within the navigation policy. Visual methods~\cite{wvn, sterling}
predict how the terrain will affect the robot from the camera image ahead. Experience-based methods~\cite{badgr, wayfast, terrapn} learn
that outcome from the robot's own driving. Uncertainty-aware
models~\cite{evora, pietra, cahsor} raise the cost of terrain they
have not seen. Risk-aware planners~\cite{step} optimize the tail of
the cost distribution, and learned policies score sampled
trajectories with a critic~\cite{navdp} or push them away from
estimated obstacles~\cite{care}. In every case, the safety issue is
addressed before a path is committed to by choosing among
candidates. Once the robot is driving the chosen path, however, no
component judges whether continuing it is still safe, and even when every
candidate is hazardous, the navigation model still selects the least
hazardous one and continues to generate paths without any signal that the
situation has changed (Fig.~\ref{fig:motivation}). The navigation
model therefore leaves a gap at runtime that only an alarm can fill,
and none of these methods raises one.\looseness=-1

Runtime failure detection has recently been studied for manipulation
and vision--language--action policies~\cite{sinha, stac, fiper, safe,
faildetect}, but under bounded episodes with observable outcomes and
repeatable resets. Continuous off-road driving offers neither. A run
has no natural endpoint, and failures cannot be reproduced at will.
Simulation has advanced considerably, yet it still does not capture
the full range of robot--terrain interaction, and a real rollover or
immobilization can end a field campaign. What the logs contain
instead is the human's intervention, which preserves the approach to
a hazard while censoring the failure it prevents, so interventions
are far more plentiful than failures.

Prior work has used these interventions as supervision. LaND~\cite{land}
predicts the upcoming intervention from the current observation, and
PAAD~\cite{paad} predicts failure states along a planned path, but both
targets are late: the intervention arrives only after the human has
recognized the hazard and reacted, and the failure state comes later
still.

We instead target the moment at which the human's judgment shifts
from safe to unsafe to continue, and we call this moment the
\emph{perception onset}. The intervention does not mark this moment,
because by the time the human takes over, the robot has already been
in a state severe enough to require intervention for some time. The
human may watch the situation develop before deciding that
intervention is necessary, so a label anchored at the intervention
describes driving that has been unsafe for a while rather than
driving that is turning unsafe. The onset is a human-dependent label
rather than an objective safety boundary, but a runtime alarm
should reproduce this judgment, and the onset precedes the
intervention-based targets discussed above.

Two properties of field logs shape the alarm. Most of the recording
captures uneventful driving, and the few onsets it contains
are sparse. Discrete-time survival analysis~\cite{suresh,prognostics}
uses both kinds of record. It models the probability that the onset
falls in each upcoming interval given that it has not occurred yet.
So a segment that reaches an onset supervises when it happened and a
segment that ends without one supervises that it had not happened
yet. A CUSUM~\cite{page} then converts the predicted risk into an
alarm. It accumulates the excess over the nominal level frame by
frame, so isolated spikes decay and only a sustained rise reaches the
threshold.

CUSP (\textbf{CU}SUM-governed \textbf{S}urvival model of the
\textbf{P}erception onset) implements this in two parts. A lightweight
hazard head over frozen vision-foundation features estimates an
onset-risk score at every frame. A one-sided CUSUM then accumulates
that score into an alarm, with the threshold set from the site's own
normal driving to meet a chosen false-alarm rate. Unlike prior intervention-supervised
alarms~\cite{land, paad, gaft}, CUSP is supervised by the human's
unsafe-to-continue judgment itself and treats the onset as a
time-to-event target rather than a binary label. To our knowledge, it
is the first runtime alarm to annotate this onset separately from
the intervention and learn it as a time-to-event target.\looseness=-1

We make three contributions.
\begin{itemize}
\item We define the \emph{perception onset}, the moment at which the
human judges continued autonomy unsafe, and annotate it separately
from the intervention, which no intervention-supervised alarm has
done before.

\item We propose CUSP, which supervises a visual hazard head on the
onset with a discrete-time survival objective and turns the predicted
onset risk into alarms with a one-sided CUSUM at a false-alarm rate
set from the site's normal driving. \looseness=-1

\item We show at five unseen sites under a common target false-alarm budget
that CUSP detects 85 of 142 events against 26 for the best of nine
adapted baselines.
\end{itemize}

\section{Related Work}
\label{sec:related_work}

\begin{figure*}[!t]
\centering
\includegraphics[width=0.92\textwidth]{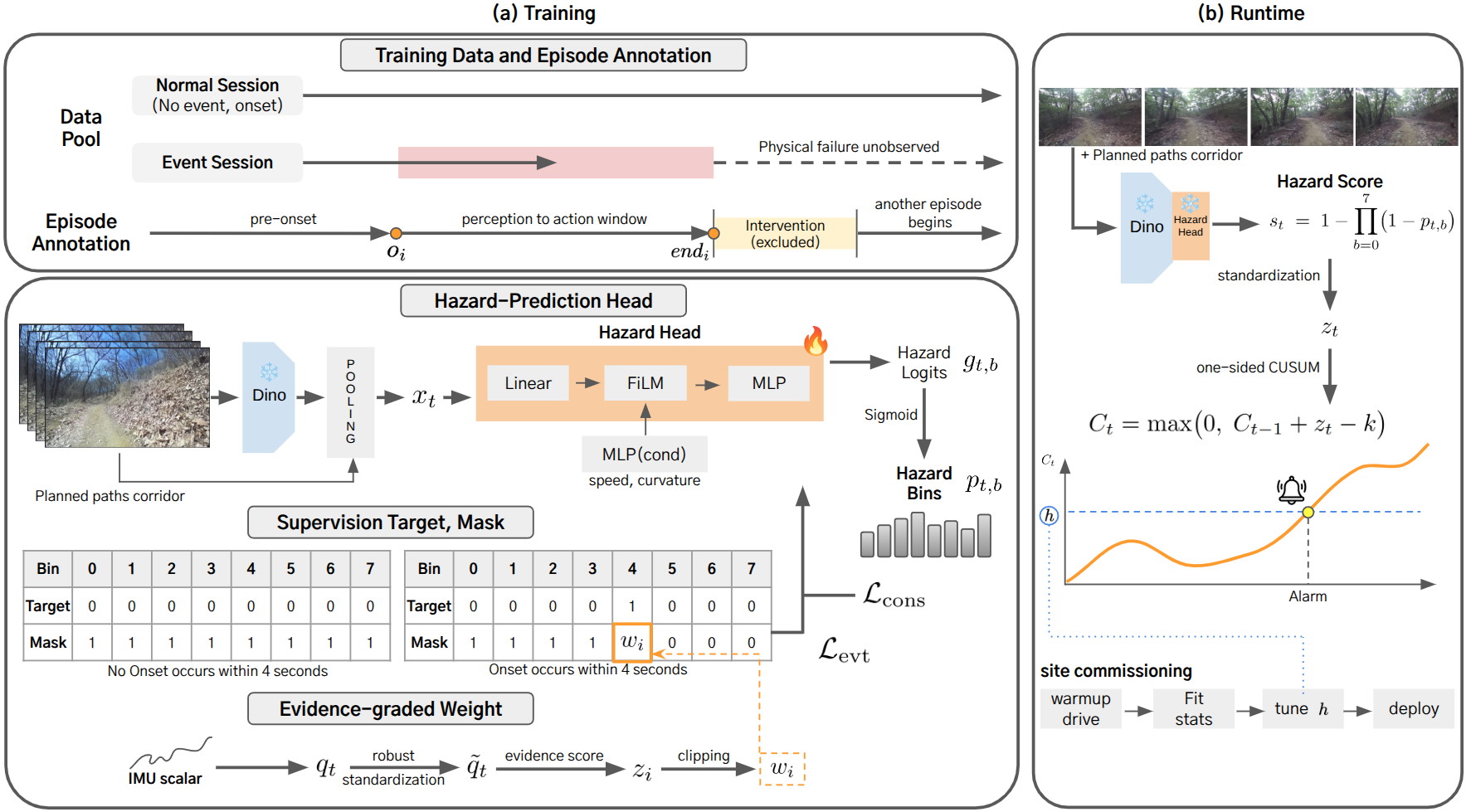}
\caption{CUSP overview. (a) Training. Each episode is annotated with the perception onset $o_i$ and the $\mathrm{end}_i$ just before intervention. Frozen DINOv3 features are pooled over the committed-path corridor with temporal differences and global context and FiLM-conditioned on path curvature and speed. Survival targets and masks are anchored on $o_i$, and per-event weights $w_i$ are graded by IMU tilt around the onset (training only). (b) Runtime. The onset-risk score is robust-standardized and accumulated by a one-sided CUSUM under a per-site false-alarm budget.}\label{fig:arch}
\end{figure*}

\subsection{Safety in Off-Road Navigation}
\label{sec:rw_offroad}
The traversability and policy methods surveyed in
Sec.~\ref{sec:introduction} decide which path to take, and none of
them judges whether the path already committed to remains safe to
continue. Vehicle-centric monitors based on zero-moment-point
criteria~\cite{zmp} and terramechanic assurance~\cite{naik2026monitor}
rest on modeled criteria and each targets one hazard mechanism, and
anomaly-based traversability estimation~\cite{wellhausen2020anomaly}
flags terrain that differs from previously traversed terrain, but
novelty and hazard do not coincide, since unfamiliar terrain may be
safe and familiar terrain may become unsafe through robot--terrain
interaction. None of these approaches uses the human's judgment as
supervision.\looseness=-1

\subsection{Supervision for Learned Runtime Alarms}
\label{sec:rw_interventions}
Runtime alarms for learned manipulation and vision--language--action
policies detect task-contextual anomalies~\cite{sinha} or flag failure
from task progress, calibration on successful rollouts, action
distributions, or internal representations~\cite{stac, fiper, safe,
faildetect} -- all on finite-horizon rollouts with an explicit end and a
success or failure label, which continuous off-road driving lacks.\looseness=-1

What off-road driving does have is the human's intervention.
LaND~\cite{land} predicts the upcoming intervention from the current
observation and candidate actions; PAAD~\cite{paad} defines binary
failure states and predicts future failure states from perception and a
planned path. GAFT~\cite{gaft} adapts a vision foundation model on
frames from intervention-verified approach runs, labeled by outcome as
binary hazard. All three anchor the label to the intervention or its outcome that occurs after the human
has already recognized the hazard. So the label is late by at least
the time the human takes to decide that intervention is necessary,
and no intervention-supervised method annotates the earlier moment at
which the human first judged continued autonomy unsafe. \looseness=-1

\subsection{Survival Analysis and Sequential Change Detection}
\label{sec:rw_stats}
Survival analysis models time to an event while retaining observations
that end without the event as right-censored information, and in
discrete time the hazard is the probability that the event occurs in an
interval given that it has not occurred earlier~\cite{suresh,
prognostics}. OSCAR~\cite{oscar} recently applied it to robot navigation
with an obstacle clearing as the event, whereas CUSP takes the
human-annotated perception onset as its event rather than the physical
outcome that intervention makes unavailable. \looseness=-1

CUSUM~\cite{page} detects sustained change by accumulating evidence for
a post-change distribution and resetting when nominal evidence
dominates, with classical optimality results under specified change
models~\cite{lorden, moustakides}. CUSP neither treats its learned score
as a likelihood ratio nor invokes these guarantees, but standardizes the score against the site's normal driving, accumulates sustained elevations, and selects the threshold empirically from that driving.

\section{Method}\label{sec:method}

CUSP attaches to an existing navigation model and requires no change
to the policy or its training. A hazard head on frozen
vision-foundation features is trained as a discrete-time survival
model on the onset (Secs.~\ref{sec:head}--\ref{sec:objective}), and
a one-sided CUSUM accumulates its runtime score under a site-local
false-alarm budget (Sec.~\ref{sec:decision}).

\subsection{Problem Definition}\label{sec:problem}

\textbf{Notation.} An \emph{intervention} is the instant the human
takes manual control, and only interventions driven by a hazard to the
platform are \emph{events}. Since intervention preempts the failure,
every event is right-censored at intervention, and CUSP predicts the onset
of human-perceived danger rather than the failure. CUSP samples the
run at rate $f$, with frame $t$ timestamped $t/f$. Frame $t$
carries the camera image $I_t$, the committed path $\mathcal{P}_t$, its
curvature $\mathrm{curv}_t$, and the platform speed $\mathrm{spd}_t$
measured from pose, which are the only runtime inputs. Event $i$ has an
\emph{onset} $o_i$, the earliest time at which continuing along
$\mathcal{P}_t$ would lead into the hazard, and an \emph{end} just
before intervention, and $[o_i,\mathrm{end}]$ is the
\emph{perception-to-action window}. The \emph{horizon} $H$ is how far
ahead the alarm looks, discretized into $B{=}H/\Delta$ bins with
$\Delta{=}1/f$. The interval $[o_i-H,o_i)$ is the \emph{positive band},
\emph{normal driving} comprises every frame outside the
perception-to-action windows and the human-controlled segments, and a
\emph{site} is one physical environment.

\textbf{Task.} An \emph{alarm} raised inside $[o_i,\mathrm{end}]$ is a
\emph{detection} of event $i$, and an alarm during normal driving is a
\emph{false alarm}, counted per accumulation episode and reported as a rate per unit of normal
driving. The \emph{false-alarm budget} $\alpha$ is the maximum such rate
a site tolerates, and it is the sole quantity the decision threshold is
tuned to meet. CUSP must detect as many onsets as possible while keeping
each site's false-alarm rate within $\alpha$, with one learned model
that is not retrained per site.

\subsection{Onset Annotation}\label{sec:Data}
Both timestamps are annotated post hoc from the recorded video.
Annotators play each session in order, mark the frame just before each
intervention as the end, then rewind and re-watch the approach to locate
the first frame at which continuing the committed path would lead into
the hazard and mark its timestamp. Three annotators marked each onset,
and the median of their timestamps defines the final label $o_i$.
This aggregation limits the influence of a single early or late timestamp.
The onset is therefore a retrospective judgment made knowing
that an intervention followed. A parallel IMU stream grades each event's label weight (Sec.~\ref{sec:grading}), and it is not used at runtime.

\subsection{Input Features and Architecture}\label{sec:head}
\textbf{Visual input.} The frozen DINOv3~\cite{dinov3} backbone encodes
$I_t$ into patch features of dimension $D{=}768$, pooled in two ways.
$\mathrm{f}^{\mathrm{plan}}_t$ is the mean over the patches that
$\mathcal{P}_t$ covers when projected into $I_t$ as a corridor of the
platform's width, and $\mathrm{f}^{\mathrm{glob}}_t$ is the mean over
all patches, so the former describes the terrain the plan is about to
enter and the latter the scene as a whole. Both use only frame $t$. As
the platform advances from frame $t$ to $t+\ell$, the corridor feature
changes by
$\delta_\ell\mathrm{f}^{\mathrm{plan}}_{t+\ell}=\mathrm{f}^{\mathrm{plan}}_{t+\ell}-\mathrm{f}^{\mathrm{plan}}_{t}$,
which is appended to the input of the later frame, so the alarm is
causal. Writing $t$ for the current frame, the visual input is
\begin{equation}\label{eq:feat}
x_t=\operatorname{concat}\!\bigl(\mathrm{f}^{\mathrm{plan}}_t,\;
\delta_{\ell_1}\mathrm{f}^{\mathrm{plan}}_t,\;
\delta_{\ell_2}\mathrm{f}^{\mathrm{plan}}_t,\;
\mathrm{f}^{\mathrm{glob}}_t\bigr)\in\mathbb{R}^{4D},
\end{equation}
where $\ell_1<\ell_2$ span about 1 and 2\,s, so the head sees what lay
ahead one and two seconds ago against what lies ahead now.

\textbf{FiLM conditioning.} The same terrain carries different risk when
driven slowly and straight versus fast through a turn, so
$\mathrm{curv}_t$ and $\mathrm{spd}_t$ modulate the visual
representation via FiLM,
\begin{equation}\label{eq:film}
\begin{aligned}
&h_0=\operatorname{LN}(W x_t),\\
&(\gamma,\beta)=\operatorname{MLP}_{\mathrm{cond}}\!\bigl([\mathrm{curv}_t,\mathrm{spd}_t]\bigr),\\
&h_1=\phi\bigl(h_0\odot(1{+}\gamma)+\beta\bigr),\\
&g_t=W_3\,\phi(W_2 h_1),
\end{aligned}
\end{equation}
applied once after the $4D\to256$ projection and layer normalization,
where $\phi$ is the SiLU activation and $W,W_2,W_3$ are linear layers.
A $256\to128\to B$ head produces the hazard logits
$g_t=(g_{t,0},\dots,g_{t,B-1})$ with $p_{t,b}=\sigma(g_{t,b})$.

\subsection{Onset Target and Evidence-Graded Supervision}\label{sec:grading}
\textbf{Hazard target.} We set $H{=}4$\,s so that the horizon exceeds
the 2.5\,s perception--reaction time adopted as a design value for a
driver facing an unexpected hazard~\cite{green2000, aashto}, leaving
a margin between the predicted risk and the human's own response.
The head predicts for each bin $b$ the conditional hazard
$p_{t,b}$ that onset occurs in bin $b$ given that it has not occurred
earlier. With $d=o_i-t/f$, a frame in the positive band
($0<d\le H$) has its onset at bin
\begin{equation}\label{eq:bo}
b_o=\min\!\bigl(\lfloor d/\Delta\rfloor,\,B{-}1\bigr)
\end{equation}
and per-bin targets and masks
\begin{equation}\label{eq:targets}
Y_{t,b}=\mathbf{1}[b=b_o]
\qquad
M_{t,b}=\begin{cases}1,& b<b_o\\ w_i,& b=b_o\\ 0,& b>b_o.\end{cases}
\end{equation}
Pre-onset bins are supervised as ``not yet,'' the onset bin is the sole
positive with weight $w_i$, and post-onset bins are masked because the
event has already occurred. A frame with no onset within the horizon has
$Y_{t,b}=0$ and $M_{t,b}=1$ for every $b$ and provides only survival
supervision. Frames in $[o_i,\mathrm{end}]$ and the human-controlled
segment after intervention are not used for training.\looseness=-1

\textbf{Evidence-graded weight $w_i$.} The onset is annotated from
video after the fact, so its timing varies with the annotator.
Following prior work that weights human-provided supervision by a
confidence score rather than using it uniformly~\cite{wu2019imperfect},
and in learning from interventions grades it by an independent
physical criterion~\cite{safehil, unarl}, we weight each onset by IMU
tilt inside the positive band. Hazards from robot--terrain
interaction act on the platform by the time the human perceives them,
so tilt shows, independently of the annotator, that the annotated
moment lies on a physical process already under way, whereas an onset
without tilt rests on the video alone. From IMU roll $\phi_t$ and pitch $\theta_t$ we form
$q_t=\sqrt{\phi_t^2+\theta_t^2}$ and robust-standardize it against
normal driving,
\begin{equation}\label{eq:zi}
\tilde{q}_t=\frac{q_t-\mathrm{med}}{c\cdot\mathrm{MAD}+\epsilon},
\end{equation}
with med and MAD over normal-driving frames, $c\approx1.4826$ the
consistency constant, and a small $\epsilon$. The evidence score is the
peak over the positive band,
\begin{equation}\label{eq:zpeak}
z_i=\max_{o_i-H\le t/f<o_i}\tilde{q}_t ,
\end{equation}
mapped to a clipped linear ramp between bounds
$z_{\mathrm{lo}}<z_{\mathrm{hi}}$,
\begin{equation}\label{eq:wi}
w_i=\operatorname{clip}\!\left(\frac{z_i-z_{\mathrm{lo}}}{z_{\mathrm{hi}}-z_{\mathrm{lo}}},\,0,\,1\right).
\end{equation}
With $z_{\mathrm{lo}}{=}1$ and $z_{\mathrm{hi}}{=}2$ an onset receives
positive weight only once its peak tilt exceeds the central spread of
normal driving and full weight at two robust standard deviations. An
event with $w_i{=}0$ is neither discarded nor relabeled as normal
driving. Its onset bin is excluded from the loss while its pre-onset
bins retain $M_{t,b}{=}1$ and $Y_{t,b}{=}0$, so the event is
right-censored at its onset and only the transition time is withheld.
The same censoring covers hazards that induce no tilt before onset,
such as a level approach to an obstacle, which the head still learns to
score since the visual change of a corridor becoming undrivable is
shared with the tilt-verified onsets on which it is trained. \looseness=-1

\subsection{Training Objective}\label{sec:objective}
\textbf{Event loss.} Positives are rare per bin, so with $\mathrm{pos}_b$
and $\mathrm{neg}_b$ the per-bin counts over the training set, the
positive term of the binary cross-entropy is scaled by
\begin{equation}\label{eq:pib}
\pi_b=\operatorname{clip}\!\left(\frac{\mathrm{neg}_b}{\max(\mathrm{pos}_b,1)},\;1,\;\pi_{\max}\right),
\end{equation}
with $\pi_{\max}$ fixed in advance, and the event loss is the
mask-normalized sum
\begin{equation}\label{eq:levt}
\mathcal{L}_{\mathrm{evt}}=\frac{\sum_{t,b}M_{t,b}\,\operatorname{BCE}_{\pi_b}(g_{t,b},Y_{t,b})}{\max\bigl(\sum_{t,b}M_{t,b},\,1\bigr)}.
\end{equation}
Under Eq.~\eqref{eq:targets} this is a weighted discrete-time survival
negative log-likelihood, in which pre-onset bins supervise
survival, the onset bin supervises the hazard, and post-onset bins leave
the risk set through the mask. \looseness=-1

\textbf{Photometric consistency.} We form $x'_t$ from
$\operatorname{aug}(I_t)$ as in Eq.~\eqref{eq:feat}, with unchanged conditioning
$c_t=[\mathrm{curv}_t,\mathrm{spd}_t]$, and
\begin{equation}\label{eq:lcons}
\mathcal{L}_{\mathrm{cons}}=\operatorname{mean}\bigl\lVert
\operatorname{head}(x_t,c_t)-\operatorname{head}(x'_t,c_t)\bigr\rVert_2^2
\end{equation}
pushes the head toward appearance invariance and stabilizes the score
scale across sites. The total objective is
$\mathcal{L}=\mathcal{L}_{\mathrm{evt}}+\lambda_{\mathrm{cons}}\mathcal{L}_{\mathrm{cons}}$.\looseness=-1

\subsection{Runtime Score and CUSUM Decision}\label{sec:decision}
\textbf{Runtime score.} The decision layer needs one scalar from the
$B$ bin-wise hazards. Each bin survives with probability $1-p_{t,b}$,
so the complement of the product is the probability of at least one
onset within the horizon,
\begin{equation}\label{eq:score}
s_t \;=\; 1-\prod_{b=0}^{B-1}\bigl(1-p_{t,b}\bigr).
\end{equation}
Because of $\pi_b$ the head is class-balanced rather than calibrated, so
$s_t$ is a ranking score and is standardized before any decision.

\textbf{CUSUM.} A hazard builds up over the approach rather than
announcing itself in a single frame. The score is standardized against
the site's normal driving as in Eq.~\eqref{eq:zi},
\begin{equation}\label{eq:zt}
z_t=\frac{s_t-\mathrm{med}_0}{c\cdot\mathrm{MAD}_0+\epsilon},
\end{equation}
and accumulated by a one-sided CUSUM~\cite{page},
\begin{equation}\label{eq:cusum}
C_t=\max\bigl(0,\;C_{t-1}+z_t-k\bigr),\qquad C_0=0.
\end{equation}
An alarm is raised at the first crossing $C_t\ge h$, after which the
statistic resets and re-arms, and one such episode is one alarm. A
frame adds to $C_t$ only when its score lies more than $k$ robust
standard deviations above the normal-driving median, so isolated
excursions decay and only sustained excess accumulates.  \looseness=-1

\textbf{Threshold and per-site commissioning.} We fix $k$ across
sites and set $h$ per site from a short unlabeled warm-up drive. The
warm-up gives $(\mathrm{med}_0,\mathrm{MAD}_0)$ and, through
Eq.~\eqref{eq:cusum}, a trace of $C_t$ on normal driving. We take $h$
as the smallest value that this trace crosses no more than $\alpha$
times per hour. Commissioning sets only
$(\mathrm{med}_0,\mathrm{MAD}_0,h)$; the head and other parameters stay
fixed.  \looseness=-1

\section{Experiments}\label{sec:results}


\subsection{Experimental Setup}\label{sec:setup}
\textbf{Platform.} All data were collected on a Clearpath Husky A200
UGV with a forward-facing ZED~2i stereo camera and built-in IMU. The
left RGB stream is the only runtime image input, and the IMU provides
the training-time evidence. The alarm runs at $f{=}2$\,Hz, and
inference takes 10\,ms per frame on a Jetson AGX Orin.  \looseness=-1

\textbf{Data.} All autonomous driving is by NoMaD~\cite{nomad}, whose
predicted trajectory serves as the committed path, with CUSP attached
as an alarm that leaves the policy's commands unchanged. Training
uses 163 intervention-terminated events and 329\,min of normal
driving from six training sites. Testing uses five held-out sites
that are physically separate from the training sites and from one
another. The 163 is the number of onsets rather than of supervision signals.
Under the survival objective all 39{,}480 image frames supervise
the head, frames in a positive band with the interval in which the
onset falls and all other frames as right-censored. Sites~I--II (11/50 events, 30.4/42.7 min of
normal driving) are driven by NoMaD, so their committed paths are
on-policy, and Sites~III--V (28/13/40 events, 9.3/49.0/73.6 min) are
teleoperated.  \looseness=-1

\textbf{Implementation details.} The backbone is a frozen DINOv3
ViT-B/16~\cite{dinov3} on $224{\times}224$ frames, and the plan corridor
samples the projected path at several lateral offsets spanning the
platform width. The hazard head is a three-layer MLP
($4D\to256\to128\to B$, $B{=}8$) with LayerNorm, SiLU, and dropout 0.1,
and a two-layer MLP produces the FiLM parameters from the z-scored pair
$[\mathrm{curv}_t,\mathrm{spd}_t]$. Photometric augmentations are
brightness, contrast, saturation, hue, gamma, and blur. Training uses
AdamW (learning rate $8{\times}10^{-4}$, weight decay
$2{\times}10^{-3}$), batch size 4096, and 70 epochs. The deployed
alarm averages five heads from independent seeds, with $\ell_1{=}2$,
$\ell_2{=}4$, $\lambda_{\mathrm{cons}}{=}1$, $\pi_{\max}{=}40$,
$\epsilon{=}10^{-9}$, and $k{=}3$ for CUSP. All
hyperparameters were fixed by leave-one-site-out cross-validation on
the training sites, and no test-site data informed any design choice.  \looseness=-1

\textbf{Kinematic conditioning.} NoMaD provides the committed path
at the training sites and Sites~I--II. The teleoperated
Sites~III--V have no such path, so we reconstruct one from the
teleoperator's realized future trajectory in the vehicle frame
using Hesai LiDAR-inertial odometry at 10\,Hz. This gives
$\mathrm{curv}_t$ as heading change per arc length, $\mathrm{spd}_t$
from pose differences clipped to $[0,3]$\,m/s, and the corridor for
$\mathrm{f}^{\mathrm{plan}}_t$. Conditioning at Sites~III--V is thus
off-policy and hindsight-based, whereas Sites~I--II use a genuine
on-policy plan.\looseness=-1

\textbf{Metrics and commissioning.} Event $i$ is detected when an
alarm is raised inside $[o_i,\mathrm{end}]$, at time $\tau_i$. We
report:
\begin{itemize}
\item \emph{Detection}: the number of events with such an alarm,
overall and per site.
\item \emph{Margin}: $\mathrm{end}-\tau_i$, reported as counts
exceeding 0.5/1/2\,s.
\item \emph{Delay}: $\tau_i-o_i$, reported as median with IQR over
detected events.
\item $C({+}3)$: the fraction of all events with delay $\le3$\,s.
\end{itemize}
All methods use the same false-alarm budget
$\alpha{=}10$\,h$^{-1}$, with thresholds set from each site's
first 2\,min of normal driving. Realized rates are measured over each site's full normal
driving.\looseness=-1

\begin{figure*}[!tb]
\centering
\includegraphics[width=0.96\textwidth]{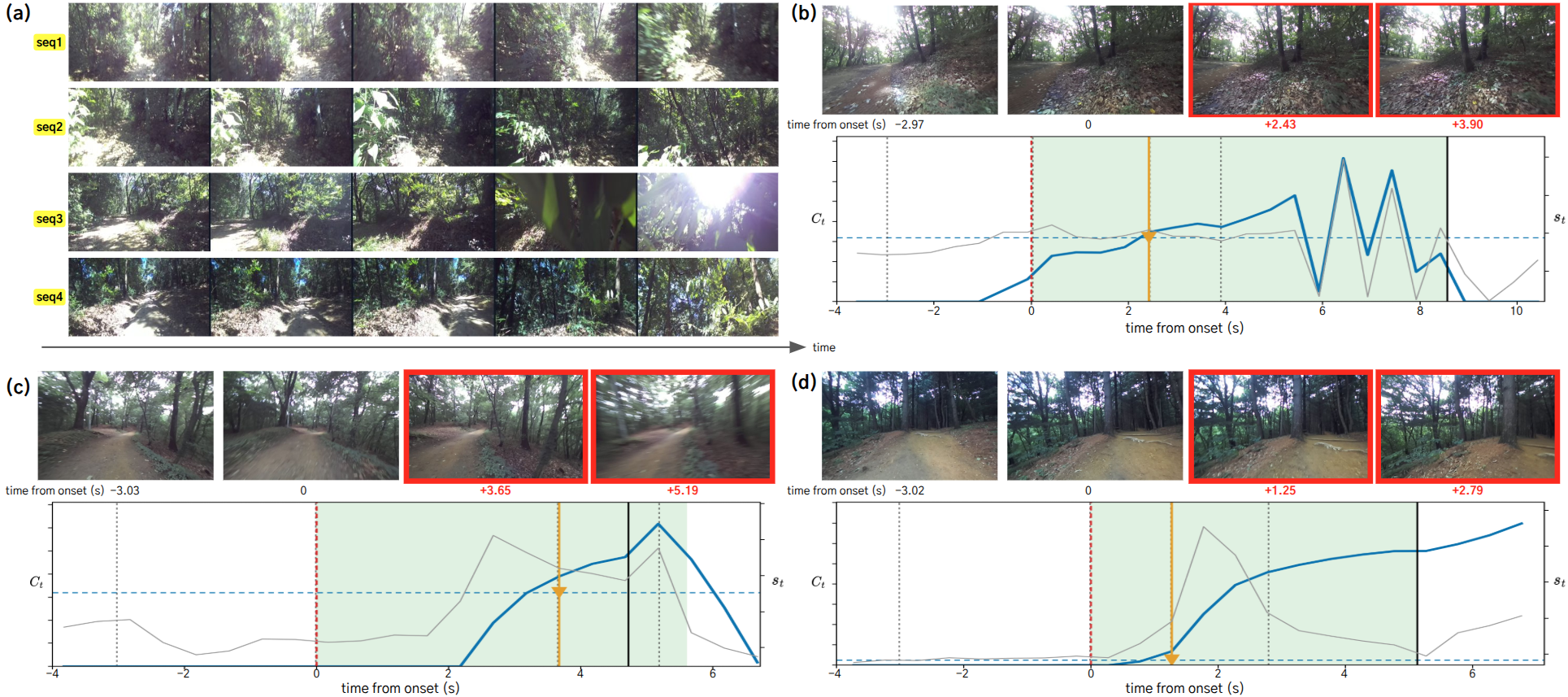}
\caption{Example detections at Sites II, III, and V. (a) Four vegetation intrusion events at Site II, where the path is visibly impassable and the baselines also detect the events. (b) Mounting a right cross-slope (Site V). (c) Sliding into a downward right cross-slope (Site III). (d) Entering a left cross-slope overgrown with dense bushes (Site III). In (b)--(d) nothing in the view marks the path as impassable, and CUSP is the only method that alarms. Each plot shows the score $s_t$ (grey) and the CUSUM statistic $C_t$ (blue) against time from onset, with the threshold $h$ as a dashed line, the perception-to-action window $[o_i,\mathrm{end}]$ shaded in green, and the alarm time marked by the orange vertical line. Frames above each plot are sampled at the indicated times from onset.}\label{fig:qual}
\end{figure*}

\textbf{Baselines.} We compare using nine methods from three
families (Table~\ref{tab:baselines}). Each method's final output is mapped to a per-frame hazard score
with the method itself left unchanged. Every learned baseline is trained or fit on the training sites and,
like CUSP, averages five heads from independent seeds; NavDP and
CARE use their released weights, and STERLING and CAHSOR, which
output a representation rather than a score, receive a two-layer
head fit to roughness. LaND, PAAD, and FIPER run on-policy at Sites~I--II; at
Sites~III--V, LaND and PAAD take the same hindsight conditioning as
CUSP, and FIPER runs the policy open-loop in goal-masked exploration
mode, as no goal is recorded. LaND and PAAD output probabilities near
zero on most normal-driving frames (93\% and 72\%), which cannot form
the tail an accumulating rule requires, so they are logit-transformed
before standardization and each baseline is reported under the better
of its raw and transformed score.  \looseness=-1

Downstream of the hazard score, every method uses robust
standardization against the site's normal driving and a threshold
commissioned per method under the same target false-alarm budget.
We compare CUSUM with $k\in\{1,2,3\}$ and the instantaneous rule per
baseline; $k{=}3$ is already the best setting for six of the nine,
including four of the five best-performing, and no baseline exceeds 26
detections under any setting, so every baseline is reported at its best
configuration.  \looseness=-1

\begin{table}[!tb]
\caption{Baselines: native output $\rightarrow$ per-frame hazard score.}
\label{tab:baselines}
\centering\scriptsize
\setlength{\tabcolsep}{3pt}
\begin{tabular*}{\columnwidth}{@{\extracolsep{\fill}}lll@{}}
\toprule
Method & Native output &  Hazard score\\
\midrule
\multicolumn{3}{@{}l}{\emph{Traversability}}\\
STERLING~\cite{sterling} & terrain feature & roughness head\\
CAHSOR~\cite{cahsor} & terrain feature & roughness head (speed-cond.)\\
SALON~\cite{salon} & visual feature & Mahalanobis to normal\\
WVN~\cite{wvn} & roughness, confidence & roughness $+$ RND novelty\\
\midrule
\multicolumn{3}{@{}l}{\emph{Obstacle proximity}}\\
NavDP~\cite{navdp} & trajectory critic & negated critic of worst sample\\
CARE~\cite{care} & repulsive force & peak force along plan\\
\midrule
\multicolumn{3}{@{}l}{\emph{Plan/policy-conditioned}}\\
LaND~\cite{land} & intervention prob. & logit\\
PAAD~\cite{paad} & failure prob. along plan & logit\\
FIPER~\cite{fiper} & novelty, action spread & conformal p-value\\
\bottomrule
\end{tabular*}
\end{table}

\begin{table}[!tb]
\caption{Five-site results under a common 10/h budget.}
\label{tab:main}
\centering\scriptsize
\setlength{\tabcolsep}{2pt}
\resizebox{\columnwidth}{!}{%
\begin{tabular}{@{}lcccc@{}}
\toprule
Method & det. ($\uparrow$)/142 (I--V) & margin ($\uparrow$)$>$0.5/1/2\,(s) & $C({+}3)$ ($\uparrow$) & delay ($\downarrow$) [IQR] (s)\\
\emph{events} & 142 (11/50/28/13/40) & & & \\
\midrule
NavDP~\cite{navdp} & 3 (0/3/0/0/0) & 3/2/2 & 0.014 & $+$2.04 [1.8--5.7]\\
CARE~\cite{care} & 10 (0/5/1/1/3) & 10/10/10 & 0.021 & $+$4.59 [1.7--5.7]\\
\midrule
LaND~\cite{land} & 12 (0/0/3/3/6) & 11/11/10 & 0.049 & $+$2.58 [1.4--3.8]\\
FIPER~\cite{fiper} & 21 (1/10/6/0/4) & 19/16/12 & 0.028 & $+$5.24 [3.2--8.4]\\
PAAD~\cite{paad} & 26 (0/14/3/1/8) & 25/22/19 & 0.056 & $+$5.13 [2.8--7.7]\\
\midrule
STERLING~\cite{sterling} & 7 (1/5/0/1/0) & 7/6/5 & 0.000 & $+$5.21 [4.4--6.0]\\
CAHSOR~\cite{cahsor} & 9 (2/6/1/0/0) & 8/7/6 & 0.014 & $+$5.60 [3.4--9.4]\\
SALON~\cite{salon} & 24 (0/20/2/2/0) & 19/17/13 & 0.028 & $+$4.69 [3.2--8.4]\\
WVN~\cite{wvn} & 26 (1/16/2/3/4) & 21/20/17 & 0.063 & $+$5.00 [2.5--7.4]\\
\midrule
CUSP (\emph{intervention label}) & 34 (3/17/8/4/2) & 34/32/24 & 0.127 & $+$2.91 [1.6--5.4]\\
\textbf{CUSP (ours)} & \textbf{85} (4/40/16/4/21) & \textbf{77/72/64} & \textbf{0.218} & $+$3.72 [1.9--6.5]\\
\bottomrule
\end{tabular}}
\end{table}

\begin{figure}[!tb]
\centering
\includegraphics[width=0.86\columnwidth]{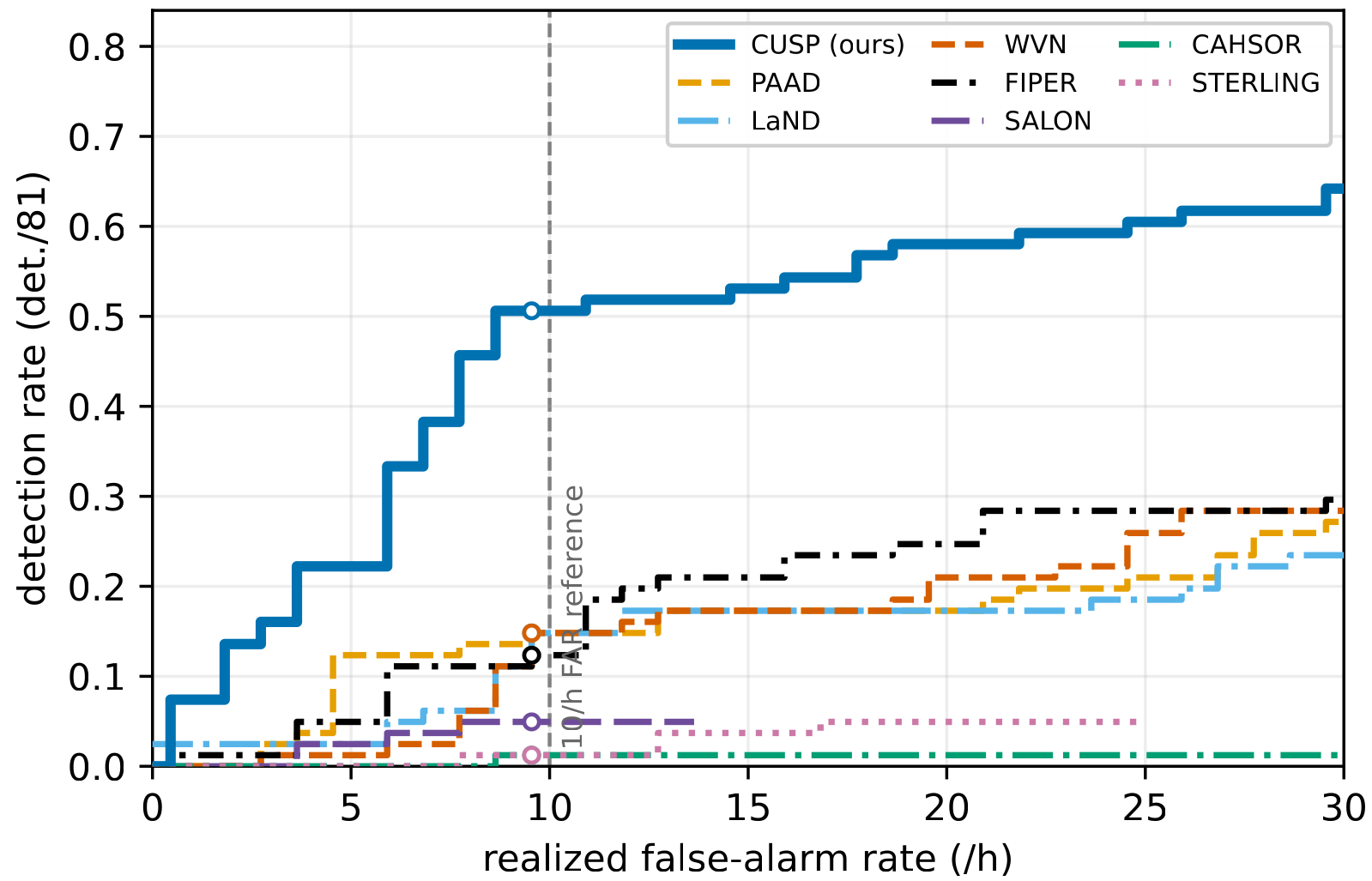}
\caption{Detection rate against realized FAR on Sites III--V as $h$ varies. The dashed line marks the 10/h operating point.}
\label{fig:budget}
\end{figure}

\subsection{Results}\label{sec:mainresults}
Table~\ref{tab:main} reports all five sites under a common 10/h budget, realized at 7.9/h at Site~I, 9.8/h at Site~II, and 9.5/h on Sites~III--V pooled (Table~\ref{tab:bakeoff}). CUSP detects 85 of 142 events, 3.3 times the 26 of the two strongest baselines, and leads at every site. The lead holds under both autonomous and teleoperated driving: on the on-policy Sites~I--II, where
CUSP ran on top of NoMaD while leaving its commands unchanged,
it detects 44 of 61 events against 20 for the best baseline, and on
the teleoperated Sites~III--V 41 of 81 against 12. It also detects
the largest fraction within 3\,s of onset: 22\% of all events against
at most 6.3\% for any baseline, and 64 of the 85 detections leave more
than 2\,s before intervention.  \looseness=-1

The detections split by hazard type. Where the hazard is a visible
obstruction, as at Site~II where the policy drives into bushes and tree trunks
that fill the corridor, the baselines also raise alarms and concentrate their
detections there. However, CUSP still detects 40 of 50 events against 20
for the best of them (Fig.~\ref{fig:qual}(a)). WVN, which combines
roughness with novelty, detects 26 events overall compared with 7 for
STERLING. At the other sites the path ahead looks
drivable, baseline detections are fewer, and the danger comes
from cross-slopes and banks acting on the platform. The question is
when continuing becomes unsafe rather than whether the path is
passable. The margin is widest (Fig.~\ref{fig:qual}(b)--(d)), and
the lead holds at every budget on Sites~III--V (Fig.~\ref{fig:budget}):
the detection rate the best baseline reaches at the 10/h budget, CUSP
reaches at under 3/h, and the 0.30 the best baseline reaches at 30/h,
CUSP reaches at 6/h. \looseness=-1

\subsection{Ablation Studies}\label{sec:ablation}

For detection-count differences (variant minus CUSP), we report
95\% percentile CIs from 4{,}000 site-stratified paired bootstrap
replicates, resampling events with replacement within each site
while preserving paired predictions. Each replicate retains the
observed event count per site and recomputes the pooled detection-count
difference. Bounds are the 2.5th and 97.5th percentiles. \looseness=-1

\textbf{Intervention label (Table~\ref{tab:main}).}
Training the same head on intervention instead of onset lowers
detection, reflecting the different positive classes. Anchored at the
intervention, the positive band $[\mathrm{end}{-}H,\mathrm{end})$
consists of frames already inside the hazard and separates from
normal driving far more cleanly than the onset band $[o_i{-}H,o_i)$.
So, the head learns to output scores near one, whereas the
onset-trained head remains moderate even inside events. Neither head
avoids occasional false outputs on normal driving, and the two
differ not in how often such outputs occur but in how large they
are. The intervention-trained head produces them at the magnitude
it learned for true events, whereas the onset-trained head, never
trained to be confident, produces small ones. Standardization turns
each large false output into a very large $z_t$, which the CUSUM
holds above $h$ for many frames, so a handful of false outputs uses
up much of the false-alarm budget and fewer events are detected at
the common budget. The onset label trains a harder class, and in
return it bounds false outputs so that the alarm releases between
events.  \looseness=-1

\begin{table}[!tb]
\caption{Training objective ablation.}
\label{tab:loss}
\centering\scriptsize
\setlength{\tabcolsep}{2pt}
\resizebox{\columnwidth}{!}{%
\begin{tabular}{@{}lccclc@{}}
\toprule
Objective & Hor. & Mask & $w_i$ & det./142 (I--V) & $\Delta$ [95\% CI]\\
\midrule
Binary~\cite{land,paad} & $H$ & --- & --- & 9 (2/1/3/0/3) & \ci{-76}{-89}{-64}\\
Per-bin binary & $B$ bins & no & --- & 4 (0/1/1/0/2) & \ci{-81}{-92}{-70}\\
Time-to-onset regr. & cont. & --- & --- & 11 (0/3/2/1/5) & \ci{-74}{-85}{-63}\\
Survival & $B$ bins & yes & uniform & 45 (0/8/15/3/19) & \ci{-40}{-49}{-32}\\
\midrule
\textbf{Survival (CUSP)} & $B$ bins & yes & graded & \textbf{85} (4/40/16/4/21) & ---\\
\bottomrule
\end{tabular}}
\end{table}

\textbf{Training objective (Table~\ref{tab:loss}).} With features,
head, consistency term, and decision layer fixed, we isolate the choice
of supervision. Binary classification~\cite{land,paad} detects 9
events, while per-bin classification and time-to-onset regression
detect 4 and 11. The survival mask (Eq.~\eqref{eq:targets}) raises
detection to 45, and evidence grading to 85. These comparisons support
modeling onset as a time-to-event target rather than classifying the
same visual inputs.  \looseness=-1

\begin{table}[!tb]
\caption{Positive label extended to $[o_i,\,o_i{+}\eta]$.}
\label{tab:window}
\centering\scriptsize
\setlength{\tabcolsep}{2pt}
\begin{tabular*}{\columnwidth}{@{\extracolsep{\fill}}lccccccc@{}}
\toprule
$\eta$ (s) & \textbf{0} & $+$1 & $+$2 & $+$4 & $+$6 & $+$10 & end\\
\midrule
det./142 & \textbf{85} & 72 & 72 & 69 & 62 & 60 & 58\\
I/II & 4/40 & 0/31 & 1/33 & 3/30 & 2/30 & 2/29 & 2/28\\
III/IV/V & 16/4/21 & 16/5/20 & 15/4/19 & 13/4/19 & 9/4/17 & 10/4/15 & 8/5/15\\
\bottomrule
\end{tabular*}
\end{table}

\textbf{Positive-bin extent (Table~\ref{tab:window}).} Supervising the
transition is more effective than supervising an ongoing hazard.
Extending the positive label to $[o_i,\,o_i{+}\eta]$, capped at event
end, lowers detection at every offset, from 85 to 72 at $\eta{=}1$\,s
and to 58 at event end, with the largest losses at Sites~II and III, where the
perception-to-action windows are longest.  \looseness=-1

\begin{table}[!tb]
\caption{Evidence-weight mapping $z_i\!\to\!w_i$.}
\label{tab:evidence}
\centering\scriptsize
\setlength{\tabcolsep}{2pt}
\resizebox{\columnwidth}{!}{%
\begin{tabular}{@{}llclc@{}}
\toprule
Scheme & $w_i$ & $w_i{>}0$/163 & det./142 (I--V) & $\Delta$ [95\% CI]\\
\midrule
Soft ramp & $\mathrm{clip}(z_i/2,0,1)$ & 146 & 41 (0/3/15/5/18) & \ci{-44}{-51}{-36}\\
Uniform (evidence off) & $1$ & 163 & 45 (0/8/15/3/19) & \ci{-40}{-49}{-32}\\
Sigmoid & $\sigma(z_i{-}1)$ & 163 & 48 (0/9/15/4/20) & \ci{-37}{-46}{-28}\\
Gate & $\mathbf{1}[z_i{\ge}1.5]$ & 89 & 62 (2/23/14/5/18) & \ci{-23}{-32}{-14}\\
\midrule
\textbf{Clipped ramp (CUSP)} & Eq.~\eqref{eq:wi}, $(1,2)$ & 114 & \textbf{85} (4/40/16/4/21) & ---\\
\bottomrule
\end{tabular}}
\end{table}

\textbf{Evidence-weight shape (Table~\ref{tab:evidence}).} The decisive
choice is censoring, giving zero weight to onsets without tilt
evidence. Keeping all or nearly all onsets detects 41--48 events; gating raises
this to 62, and grading with the clipped ramp to 85. Censoring does
not raise the score on average but sharpens its upper tail, and only
frames more than $k$ robust standard deviations above the normal-driving median increase
$C_t$. At Site~II the events accumulated to about half of $h$ under
uniform weighting and to about twice $h$ under the clipped ramp, so
the sharper tail moved most of them across the threshold. The largest
sensitivity to the mapping occurs at Site~II, where hazards are visible
obstructions on level ground, so censoring tilt-less onsets does not
sacrifice that hazard type.  \looseness=-1

\begin{table}[!tb]
\caption{Decision statistic on the frozen CUSP score. Site-$\sigma$ is the population SD of the five FARs.}
\label{tab:bakeoff}
\centering\scriptsize
\setlength{\tabcolsep}{1.5pt}
\resizebox{\columnwidth}{!}{%
\begin{tabular}{@{}llccc@{}}
\toprule
Decision statistic & det./142 (I--V) & $\Delta$ [95\% CI] & site-$\sigma$ (/h) & FAR (/h, I--V)\\
\midrule
Conformal martingale & 16 (1/3/4/5/3) & \ci{-69}{-81}{-57} & 5.9 & 5.9/8.4/6.4/19.6/2.4\\
Sliding sum & 26 (2/2/8/7/7) & \ci{-59}{-69}{-48} & 3.3 & 7.9/9.8/12.9/14.7/5.7\\
EWMA & 37 (2/12/8/6/9) & \ci{-48}{-59}{-38} & 4.0 & 13.8/9.8/6.4/15.9/5.7\\
Instantaneous & 53 (2/20/10/6/15) & \ci{-32}{-43}{-22} & 6.4 & 17.8/9.8/0.0/15.9/6.5\\
\midrule
\textbf{CUSUM (CUSP)} & \textbf{85} (4/40/16/4/21) & --- & \textbf{2.2} & 7.9/9.8/12.9/12.2/7.3\\
\bottomrule
\end{tabular}}
\end{table}

\begin{table}[!tb]
\caption{Kinematic conditioning on autonomous Sites I--II.}
\label{tab:film-autonomous}
\centering\scriptsize
\setlength{\tabcolsep}{2pt}
\resizebox{\columnwidth}{!}{%
\begin{tabular}{@{}lccc@{}}
\toprule
Conditioning & det./61 (I/II) & $\Delta$ [95\% CI] & FAR (/h) [95\% upper]\\
\midrule
FiLM off & 8 (0/8) & \ci{-36}{-43}{-28} & 8.2 [$\le$18.2]\\
concat.\ $[\mathrm{curv},\mathrm{spd}]$ & 36 (0/36) & \ci{-8}{-14}{-3} & 8.2 [$\le$19.5]\\
\midrule
\textbf{FiLM (CUSP)} & \textbf{44} (4/40) & --- & 9.0 [$\le$19.5]\\
\bottomrule
\end{tabular}}
\end{table}

\textbf{Decision statistic (Table~\ref{tab:bakeoff}).} All five rules
run on the same $z_t$ under the same commissioning. The instantaneous
statistic thresholds $z_t$ directly, the sliding sum adds $z_t$ over
$W{=}B$ frames as in FIPER~\cite{fiper}, the EWMA uses $\lambda{=}0.3$,
and the conformal martingale~\cite{vovk2003exch} thresholds the power
martingale of $z_t$ p-values with $\varepsilon{=}0.5$. CUSUM detects
the most events because it treats normal driving and hazards
differently. On normal driving about 97\% of frames lie below the drift
$k$ and do not increase $C_t$; negative increments reduce accumulated
evidence toward zero. On a hazard the score rises
weakly but persistently over several seconds, and accumulation turns
that persistence into amplitude, so 60\% of events reach $h$ against
37\% for the memoryless instantaneous statistic. CUSUM also has the
lowest between-site dispersion of realized FARs, whereas the
instantaneous and EWMA statistics exceed the budget at Sites~I and IV and fall
below budget at Sites~III and~V. \looseness=-1

\textbf{Kinematic conditioning (Table~\ref{tab:film-autonomous}).} On
the autonomous Sites~I--II, with each variant re-commissioned under the
same budget, removing FiLM detects 8 events and plain concatenation of
$[\mathrm{curv}_t,\mathrm{spd}_t]$ detects 36, against 44 for FiLM at a
comparable realized rate. Neither alternative detects any events at Site~I. Speed and curvature
therefore help most when they modulate the visual representation. \looseness=-1

\section{Conclusion and Future Work}\label{sec:conclusion}
We introduced CUSP, a runtime hazard alarm for off-road navigation.
We treat the human-perceived onset, annotated separately from
intervention, as a time-to-event target. A discrete-time survival model
learns from observed onsets and driving without onsets in
intervention-terminated logs, and CUSUM converts onset risk into alarms calibrated from
normal driving. CUSP requires no failure data or policy changes.
Across five unseen sites under a common false-alarm budget, it
detected 85 of 142 events against 26 for the best of nine adapted
baselines, with the largest gains on hazards that arise from driving
the terrain rather than from a visible obstruction.  \looseness=-1

\textbf{Limitations and future work.} Three limitations point to the
next steps. First, what CUSP can detect is bounded by the hazard types
the training onsets span. Coverage saturates at about 60\% of the
events on the teleoperated Sites~III--V, where raising the budget only adds false alarms
(Fig.~\ref{fig:budget}), and at about 80\% on Site~II, so the
remaining misses may reflect hazards underrepresented in training.
Second, the alarm sees only what the low, narrow
platform-mounted camera sees, which leaves a ditch behind a crest or a
berm below the near horizon without image evidence until the platform
is on it, so a wider or elevated field of view is the next step. Third,
CUSP is only a detector and leaves recovery, slowing, or re-planning to
the human or the navigation model; the human's actions inside the recorded
perception-to-action window are a natural supervision source for that
response, and learning it from the same logs is the next step.  \looseness=-1

\end{document}